\pdfoutput=1
\documentclass[10pt,twocolumn,letterpaper]{article}
\usepackage[margin=0.75in,columnsep=0.24in]{geometry}
\usepackage[hyphens]{url}
\usepackage{graphicx}
\usepackage{natbib}
\usepackage{caption}
\usepackage{amsmath,amssymb,amsfonts}
\usepackage{booktabs}
\usepackage{multirow}
\usepackage[table]{xcolor}
\usepackage[colorlinks=true,allcolors=blue]{hyperref}

\title{I2VShield: An Efficient Proactive Defense Framework against DiT-based~Image-to-Video~Models}
\author{
Yimao Guo\textsuperscript{1} \quad
Zuomin Qu\textsuperscript{2} \quad
Wei Lu\textsuperscript{1}\\[0.5em]
\textsuperscript{1}School of Computer Science and Engineering, Sun Yat-sen University\\
Ministry of Education Key Laboratory of Information Technology\\
Guangdong Province Key Laboratory of Information Security Technology\\
Guangzhou 510006, China\\
\texttt{guoym39@mail2.sysu.edu.cn} \quad
\texttt{luwei3@mail.sysu.edu.cn}\\[0.5em]
\textsuperscript{2}State Key Laboratory of HVDC,\\
China Southern Power Grid Electric Power Research Institute\\
\texttt{quzuomin@csg.cn}
}
\date{}

\begin{document}
			
			\maketitle
			\begin{abstract}
				
				The rapid advancement of video generation models has led to the increasing misuse of image-to-video (I2V) models. Although substantial progress has been made in detecting AI-generated videos, proactive defenses against I2V models remain underexplored. In particular, current proactive defenses against I2V models predominantly rely on gradient-based adversarial attacks, which require defenders to possess GPUs with substantial memory resources (VRAM) to generate adversarial examples. To address this issue, we propose I2VShield, a privacy protection method based on generative adversarial attacks tailored to Diffusion Transformer (DiT)-based I2V models. The proposed method primarily consists of two components: (1) a text-adaptive perturbation generation framework integrating adversarial learning to mitigate computational overhead while maintaining visual imperceptibility; and (2) an untargeted Multimodal Attention Disruption (MAD) attack that exploits the inherent vulnerabilities of DiT-based I2V models, maximizing the deviation of the internal attention features from their clean states. Extensive experiments demonstrate that our approach achieves highly competitive protection performance across various datasets and mainstream DiT-based I2V models, particularly in disrupting spatiotemporal coherence, while substantially reducing computational costs.
				
			\end{abstract}

			\section{Introduction}
			
			The rapid evolution of generative artificial intelligence has catalyzed unprecedented advancements in image-to-video (I2V) generation. Driven by the recent paradigm shift toward Diffusion Transformers (DiT), modern I2V models exhibit remarkable capabilities in generating temporally coherent and highly realistic video sequences from static reference images and textual prompts \cite{hong2022cogvideo}. However, the democratization and widespread availability of these powerful I2V tools have inevitably led to rampant abuse. Malicious actors frequently exploit these models to animate private portraits without authorization, generating deepfakes and malicious content that pose serious threats to personal privacy and public safety.

			\begin{figure}[!htbp] 
				\centering
				\includegraphics[width=\linewidth]{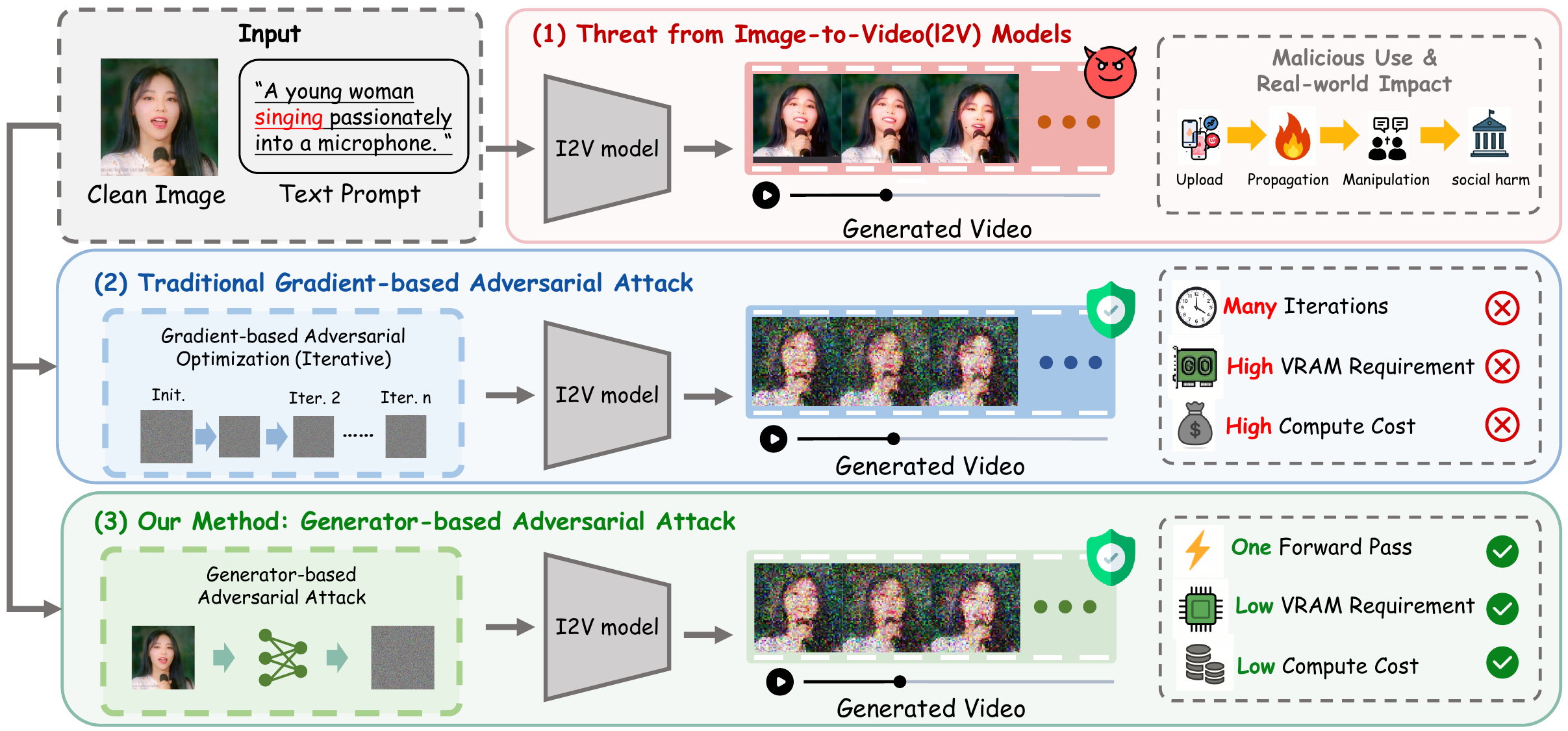}
				\caption{Background and motivation for proactive defense. (1) The abuse of I2V models poses severe privacy threats. (2) Traditional gradient-based adversarial attacks rely on expensive iterative processes, incurring high computational overhead and massive VRAM requirements. (3) Our proposed I2VShield utilizes an efficient generative attack framework to achieve effective defense at a low computational cost.}
				\label{Overview}
			\end{figure}

			To mitigate the negative impacts of generative models, extensive research has been dedicated to AI-generated content (AIGC) forgery detection \cite{rossler2019faceforensics++, wang2020cnn, xu2025multimodal}. While these passive detection mechanisms have achieved notable progress, they are inherently reactive; they can only identify malicious content after it has been generated and potentially disseminated. Consequently, there is an urgent need for proactive defense strategies that prevent the unauthorized generation of videos from the outset.  
			Adversarial perturbations have emerged as a promising proactive defense mechanism that protects users' privacy by modifying personal images before they are shared online.
			When an I2V model attempts to process these protected images, the perturbation disrupts the generative process, resulting in severe degradation or semantic collapse in the video.

			Despite the conceptual viability of proactive defenses, applying existing adversarial attack methods to modern DiT-based I2V models presents significant challenges. Current proactive defenses predominantly rely on iterative, gradient-based optimization techniques (e.g., PGD-based attacks) to craft adversarial examples~\cite{qu2025idguard}. While effective for image-to-image tasks, extending these gradient-based attacks to the video domain is highly problematic. Generating perturbations for I2V models requires backpropagating gradients through the complex diffusion process across multiple video frames. This optimization process demands prohibitive computational resources and substantial GPU memory, rendering it impractical for everyday users or defenders with limited hardware budgets. Furthermore, existing methods often fail to exploit the specific architectural vulnerabilities inherent to Diffusion Transformers, resulting in limited protection effectiveness against state-of-the-art I2V models.

			To address these critical limitations, we propose I2VShield. To the best of our knowledge, this work introduces the {first generative adversarial attack framework} specifically tailored for DiT-based I2V models. 
			Unlike prior instance-wise optimization methods, I2VShield employs a text-adaptive perturbation generation network that predicts adversarial perturbations in a single forward pass.
			This design eliminates iterative gradient computation during deployment, substantially reducing computational  requirements.
%			This design essentially eliminates the need for heavy gradient calculations during the protection phase, drastically reducing the computational overhead and VRAM requirements for end-users.

			Furthermore, to maximize the disruptive capability of the generated noise, we introduce an untargeted Multimodal Attention Disruption (MAD) attack. By analyzing the conditioning mechanisms of DiT architectures, 
%			we find that the cross-attention modules are highly vulnerable. 
			we find that cross-attention representations provide an effective optimization target for disrupting multimodal conditioning.
			Our attack explicitly targets these multimodal attention mechanisms within the latent space, maximizing the discrepancy between the original and adversarial features.

			The main contributions of this work are summarized as follows:
			
			\begin{itemize}
				
                \item We formulate proactive privacy protection for DiT-based I2V generation and propose I2VShield, a generative adversarial framework that protects reference images before they are used for unauthorized video synthesis.
            
                \item We develop a text-adaptive perturbation generator together with an untargeted Multimodal Attention Disruption (MAD) objective, enabling prompt-conditioned, $\ell_{\infty}$-bounded perturbation generation in a single forward pass while directly disrupting the multimodal conditioning pathway of DiT-based I2V models.
            
                \item We conduct extensive experiments on representative portrait and action-video benchmarks across multiple DiT-based I2V models, demonstrating superior protection effectiveness with substantially lower computational cost than gradient-based defenses.
				
			\end{itemize}

			\section{Related Work}

			\subsection{I2V Generation Models}
			
			Early video generation methods mainly relied on Generative Adversarial Networks (GANs) and autoregressive models, but often struggled with long-term temporal consistency and high-resolution synthesis \cite{tulyakov2018mocogan, yan2021videogpt}. Diffusion models (DMs) subsequently improved generation quality and training stability \cite{ho2020denoising, rombach2022high}. Models such as Video LDM and AnimateDiff extended pretrained text-to-image U-Nets with temporal attention or convolution layers to promote frame-to-frame coherence \cite{blattmann2023align, guo2023animatediff}. These architectures combine strong image-generation priors with explicit temporal modeling.

			More recently, 
%			Diffusion Transformers (DiTs) have replaced U-Net backbones with scalable Transformer architectures. 
			Diffusion Transformers (DiTs) have increasingly been adopted as scalable alternatives to U-Net backbones. 
			DiT-based systems such as Sora and Latte achieve strong physical realism and temporal consistency \cite{peebles2023scalable, brooks2024video, ma2024latte}. Modern I2V models align text prompts with reference images through latent self- and cross-attention. This dependence on multimodal feature alignment also creates a vulnerability: small disruptions to the conditioning pathway can substantially degrade the generated video, which motivates our approach.

			\subsection{Proactive Defenses and Adversarial Attacks}

Proactive defenses protect visual content from unauthorized generative-AI processing by adding imperceptible adversarial perturbations. A series of studies focuse on preventing unauthorized customization. Glaze protects artistic styles from imitation, while Anti-DreamBooth disrupts subject-driven personalization of text-to-image models \cite{shan2023glaze,van2023anti}. Another series of studies target unauthorized generative editing. PhotoGuard protects images against inpainting manipulation, and I2VGuard extends such protection to I2V models by preventing reference images from being maliciously animated \cite{salman2023raising,gui2025i2vguard}. These editing-oriented defenses typically rely on iterative, gradient-based methods such as PGD to optimize perturbations for each input image. For diffusion-based I2V models, repeatedly unrolling the denoising process and backpropagating across multiple video frames incur substantial computation and memory costs. They also require repeated access to the target model and its gradients during deployment.

In contrast, I2VShield learns a text-adaptive perturbation generator that produces protected images in a single forward pass, avoiding instance-specific optimization at inference time. Once trained, it retains prompt awareness without repeatedly accessing the target I2V model. 

            \begin{figure*}[!htbp]
				\centering
				\includegraphics[width=\linewidth]{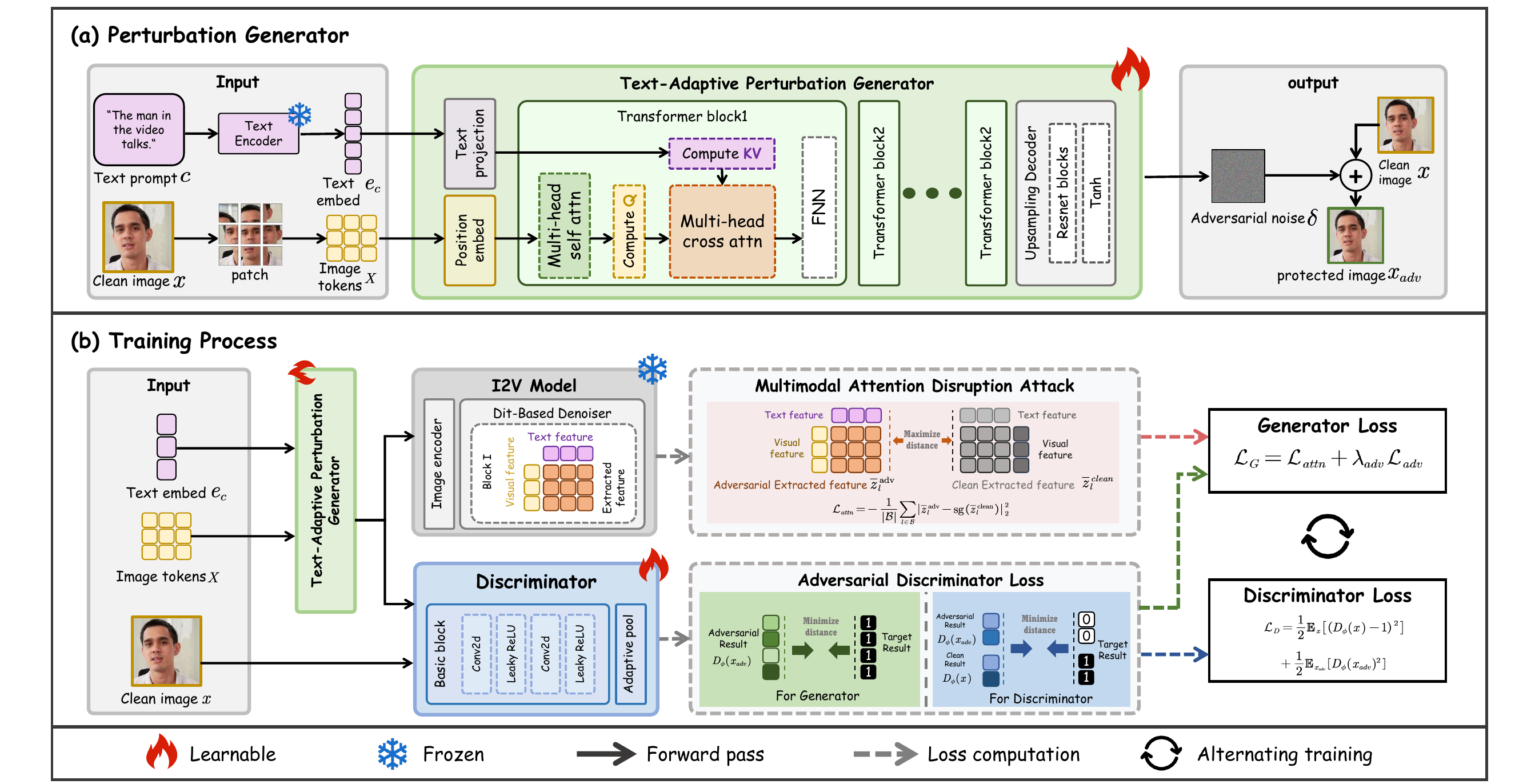}
				\caption{Illustration of the proposed I2VShield framework. (a) Perturbation generation pipeline, highlighting the architecture and key design of the Text-Adaptive Perturbation Generator. (b) Training pipeline of I2VShield, where the perturbation generator and discriminator are alternately optimized using the proposed loss functions.}
				\label{fig:method}
			\end{figure*}
            
			\section{Preliminaries}
			\subsection{Threat Model}
			\subsubsection{Attacker's Goal}
			We consider a malicious attacker who exploits an I2V generation model $\mathcal{F}$ to synthesize unauthorized content, such as deepfakes. Given a victim's reference image $x \in \mathbb{R}^{H \times W \times 3}$ and a text prompt $c$, the attacker generates a realistic and temporally coherent video $v \in \mathbb{R}^{F \times H' \times W' \times 3}$, where $F$ denotes the number of frames:
			\begin{equation}
				v = \mathcal{F}(x, c).
			\end{equation}
			\subsubsection{Defender's Capabilities}
			We assume that the defender has white-box access to the target DiT-based I2V model only during offline training. The defender uses its architecture, pretrained weights, and gradients to train the perturbation generator. 
			During online protection, the target I2V model and discriminator are no longer required; the defender encodes the prompt once and generates the protected image through a single forward pass of the perturbation generator, without accessing the target I2V model or computing its gradients.

			\subsubsection{Defender's Goal}
			Before publishing an image, the defender injects an imperceptible perturbation $\delta$ to obtain $x_{adv}=x+\delta$. When used as the visual condition of the I2V model, $x_{adv}$ should cause spatial degradation, temporal inconsistency, or semantic collapse in the generated video $v_{adv}=\mathcal{F}(x_{adv},c)$. The objective is formulated as:
			\begin{equation}
				\max_{\delta} \mathcal{D}\big(\mathcal{F}(x+\delta,c),\mathcal{F}(x,c)\big)
			\end{equation}
			\begin{equation}
				\text{s.t.} \quad \|\delta\|_\infty \leq \epsilon, \quad x+\delta \in [0,1]^{H \times W \times 3},
			\end{equation}
			where $\mathcal{D}$ measures the discrepancy between adversarial and clean generations, and the $\ell_\infty$ constraint limits the perturbation magnitude, and $\epsilon$ means the perturbation budget.
			\subsection{Defense Methods}
			\subsubsection{Traditional Methods}
			Existing proactive defenses commonly optimize $\delta$ through iterative gradient-based methods \cite{kurakin2016adversarial,madry2017towards}. For example, Iterative fast gradient sign method (I-FGSM) updates the adversarial image at step $t$ as:
			\begin{equation}
				x_{adv}^{t+1}
				=
				\text{Clip}_{x,\epsilon}
				\left(
				x_{adv}^{t}
				+
				\alpha \cdot
				\text{sgn}
				\left(
				\nabla_x
				\mathcal{D}
				\big(
				\mathcal{F}(x_{adv}^{t},c),
				\mathcal{F}(x,c)
				\big)
				\right)
				\right),
			\end{equation}
			where $\alpha$ is the step size and $\text{Clip}_{x,\epsilon}$ constrains the result within the $\epsilon$-ball around $x$. For I2V models, computing the input gradient requires backpropagating through the diffusion denoising process over multiple frames, resulting in prohibitive computation and GPU memory consumption.
			\subsubsection{Generative Defense Methods}
			To avoid instance-specific optimization at inference time, we introduce a parameterized perturbation generator $G_{\theta}$. Instead of directly optimizing $\delta$ for each input, the generator parameters $\theta$ are learned over a training dataset:
			\begin{equation}
				\max_{\theta}
				\mathbb{E}_{x,c}
				\left[
				\mathcal{D}
				\big(
				\mathcal{F}(x+G_{\theta}(x,e_c),c),
				\mathcal{F}(x,c)
				\big)
				\right],
			\end{equation}
			where $e_c$ denotes the embedding of the text prompt $c$.
			
			After offline training, $G_{\theta}$ produces an input-specific perturbation in a single forward pass. This transfers the computational cost to training and enables efficient $O(1)$ protection at inference. The key challenge is therefore to design an effective perturbation generator and suitable training objectives, as discussed in the following section.
			
			\section{Method}
			\subsection{Overview}
            An overview of I2VShield is shown in Figure~\ref{fig:method}. Existing proactive defenses iteratively optimize image-specific perturbations, incurring substantial computational overhead. Moreover, their image-only objectives overlook the joint visual-text conditioning of modern DiT-based I2V models, making prompt-agnostic perturbations potentially ineffective.

            I2VShield addresses these limitations by training a lightweight text-adaptive perturbation generator offline, enabling protection with a single forward pass. It combines an untargeted Multimodal Attention Disruption attack, which disrupts image-prompt interactions related to subject identity, semantic alignment, and temporal coherence, with discriminator-based regularization for visual fidelity. During training, the frozen I2V model processes clean and protected image-prompt pairs, while the generator maximizes their internal multimodal feature discrepancy under fidelity constraints. After training, the I2V model and discriminator are discarded, retaining only the perturbation generator and text encoder for online protection.
			
			\subsection{Framework of I2VShield}
			Existing proactive defenses typically rely on instance-wise iterative optimization, which incurs substantial computational and memory costs for I2V models with high-dimensional spatiotemporal latents. To improve deployment efficiency, I2VShield amortizes this process into a lightweight generator that produces adversarial perturbations in a single forward pass.
			
			The generator is additionally conditioned on the text prompt because different prompts may activate distinct image–text interactions for the same reference image. Incorporating prompt embeddings therefore enables the generator to produce context-adaptive perturbations that more effectively disrupt prompt-specific generation.
			
			Let $E_{\mathrm{txt}}$ denote the frozen text encoder associated with
			the target I2V model. For a prompt $c$, it produces
			\begin{equation}
				e_c = E_{\mathrm{txt}}(c),
				\qquad
				e_c \in \mathbb{R}^{L_t \times d_e},
			\end{equation}
			where $L_t$ is the number of text tokens and $d_e$ is the text-embedding
			dimension.

			At deployment time, given a clean reference image $x$ and a text prompt $c$, the prompt is first encoded into a text embedding $e_c$. The trained generator $G_{\theta}$ then predicts the perturbation and produces the protected image:
			\begin{equation}
				\delta = G_{\theta}(x, e_c), \qquad
				x_{adv} = x + \delta.
			\end{equation}
			
			The generator contains three lightweight components: image tokenization, text conditioning, and multimodal fusion with perturbation decoding.
            
			\subsubsection{Image tokenization} The input image is divided into non-overlapping patches through a convolutional patch embedding layer \cite{dosovitskiy2020image}:
			\begin{equation}	
				X_0 = \mathrm{PatchEmbed}(x) + P_{pos}, \qquad X_0 \in \mathbb{R}^{N_p \times d},
			\end{equation}
			where $N_p = (H/p)(W/p)$ is the number of patches, $p$ is the patch size, $d$ is the token dimension, and $P_{pos}$ is a learnable positional embedding. This tokenized representation allows the generator to model long-range visual dependencies while keeping the architecture compact.
			
			\subsubsection{Text conditioning} The prompt embedding is projected into the same latent dimension as the image tokens:
			\begin{equation}	
				T = \mathrm{MLP}(e_c), \qquad T \in \mathbb{R}^{L_t \times d},
			\end{equation}
			where $L_t$ denotes the number of text tokens, and MLP denotes the multilayer perceptron. This projection transforms the textual condition into a form that can be effectively fused with visual tokens.
			
			\subsubsection{Multimodal fusion and perturbation decoding} The visual tokens are processed by a stack of lightweight transformer blocks. Within each block, 
			self-attention first captures global visual context by enabling interactions between distant image regions.
			Cross-modal interaction is then introduced by using the projected text tokens as semantic guidance, enabling the visual representation to adapt to the prompt-dependent generation scenario. Finally, feed-forward layers refine the fused representation and produce feature patterns that are effective for disrupting the I2V model.
			
			After the final transformer block, the fused tokens are reshaped into a spatial feature map and passed through a cascaded upsampling decoder. The decoder predicts a dense perturbation map with the same spatial resolution as the input image. Because the entire process is feed-forward, I2VShield avoids iterative gradient computation during deployment and provides efficient proactive protection.
			
            \subsection{Multimodal Attention Disruption Attack}
			
\begin{figure}[t]
	\centering
	\includegraphics[width=\columnwidth]
	{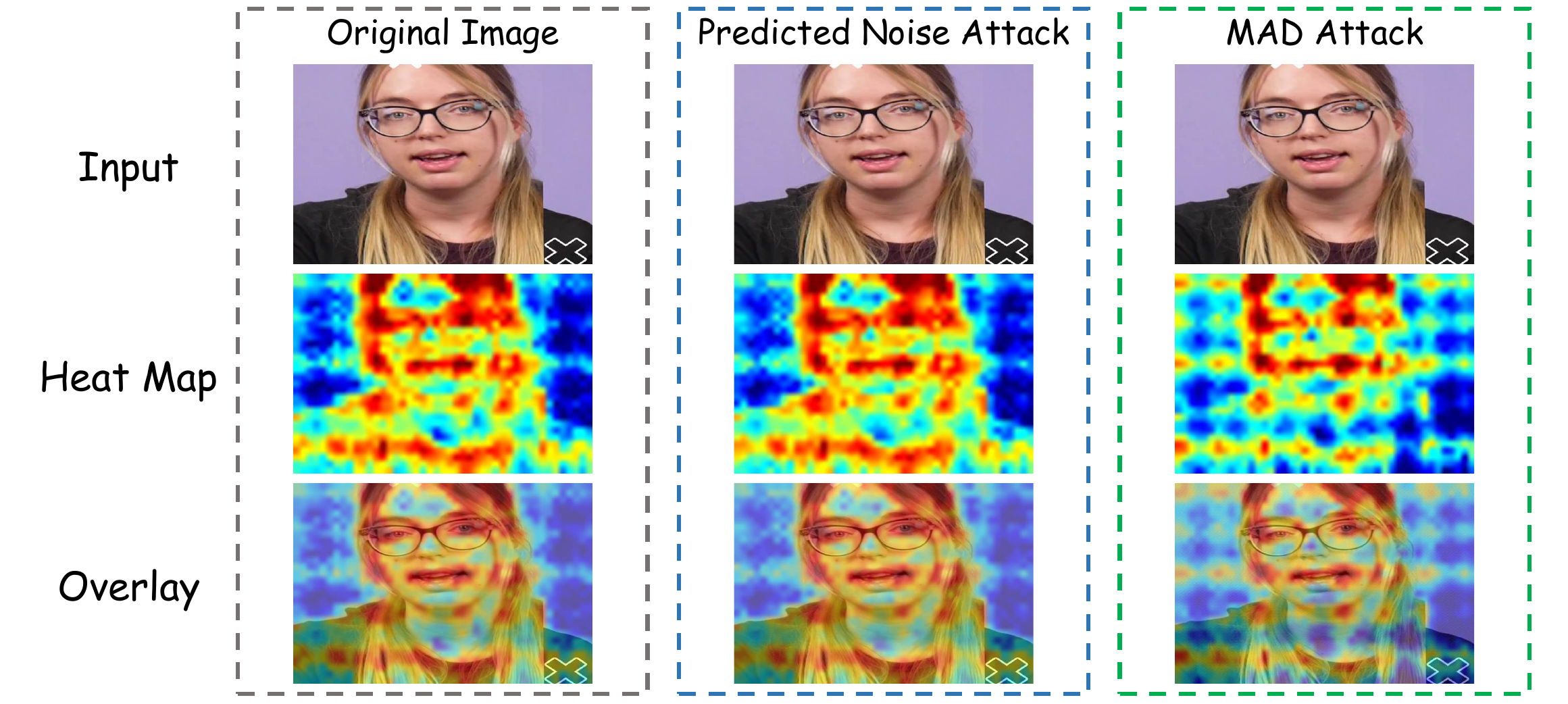}
	\caption{
		{Multimodal attention maps obtained from a clean image and images protected using the predicted noise attack and the Multimodal Attention Disruption (MAD) attack.}
	}
	\label{fig:attention_visualization}
\end{figure}

DiT-based I2V models use multimodal attention to fuse reference-image identity and appearance with textual semantic and motion cues. 
As shown in Figure~\ref{fig:attention_visualization}, we compare the attention maps of the I2V model for adversarial images generated by attacking the predicted noise and cross-attention features, respectively. 
The training objective of attacking the predicted noise is to maximize $ \left\|\mathrm{Denoiser}(x_\mathrm{adv}, c, \tau,\xi)-\mathrm{Denoiser}(x, c, \tau,\xi)\right\|_2^2 $, where $\tau$ means the timestep and $\xi$ means a noisy latent. 
Attacking the predicted noise only slightly changes the attention distribution, whereas attacking cross-attention features substantially redistributes attention away from identity-relevant regions.
This suggests that directly disrupting attention features destabilizes multimodal conditioning more effectively than perturbing noise prediction.

Motivated by this observation, we propose the untargeted Multimodal Attention Disruption (MAD) attack that maximizes the discrepancy between clean and adversarial cross-attention features. By corrupting cross-modal associations, MAD degrades subject fidelity, prompt alignment, and temporal coherence without requiring a predefined adversarial target concept or target video.

			Formally, let $\mathcal{F}$ denote the frozen DiT-based I2V model. During training, we sample a diffusion timestep 	$\tau \sim \mathcal{U}\{1,\ldots,T_d\}$ and construct the same noisy latent $\xi \sim \mathcal{N}(0,I)$ for the clean and adversarial branches. Given a selected set of transformer blocks $\mathcal{B}$, we extract the multimodal attention features $z_l$ from each block. The clean and adversarial feature states are denoted as:
			\begin{equation}	
				z_l^{\mathrm{clean}} = \Phi_l(x, c, \tau,\xi), \qquad
				z_l^{\mathrm{adv}} = \Phi_l(x_{adv}, c, \tau,\xi), \quad l \in \mathcal{B},
			\end{equation}
			where $\Phi_l(\cdot)$ denotes the multimodal attention feature extractor at the $l$-th block of $\mathcal{F}$. Both branches share the same prompt $c$, timestep $\tau$, and diffusion noise state $\xi$, ensuring that the measured discrepancy is caused by the injected perturbation rather than by stochastic variation.
			
			To reduce the influence of layer-wise scale differences, we normalize the extracted features:
%			\begin{equation}
%				\bar{z}_l =
%				\frac{z_l}{\lVert z_l \rVert_2 + \eta}
%			\end{equation}
			\begin{equation}
				\bar{z}_l^{b}
				=
				\frac{z_l^{b}}
				{\lVert z_l^{b}\rVert_2+\eta},
				\qquad
				b\in\{\mathrm{clean},\mathrm{adv}\},
			\end{equation}
			
			where $\eta$ is a small constant for numerical stability. The MAD loss is then defined as:
			\begin{equation}
				\mathcal{L}_{attn}
				=
				-\frac{1}{|\mathcal{B}|}
				\sum_{l \in \mathcal{B}}
				\left\|
				\bar{z}_l^{\mathrm{adv}}
				-
				\mathrm{sg}\left(\bar{z}_l^{\mathrm{clean}}\right)
				\right\|_2^2,
			\end{equation}

			where $\mathrm{sg}(\cdot)$ denotes the stop-gradient operation. Minimizing $\mathcal{L}_{attn}$ maximizes the feature discrepancy between the clean and protected branches. Since the clean branch is detached, the generator is encouraged to move the adversarial representation away from the original multimodal attention state without altering the I2V model itself.
			
			This untargeted objective has two advantages. First, it does not require a target video or a predefined adversarial semantic concept, which makes the training process simple and broadly applicable. Second, it attacks the feature-level alignment mechanism that is essential for DiT-based I2V generation, thereby inducing degradation in subject consistency, prompt consistency, and temporal coherence simultaneously.
			\subsection{Discriminator and Visual Fidelity Regularization}
			
			Strong feature disruption may introduce structured or high-frequency artifacts that remain noticeable even under an $\ell_{\infty}$ constraint, reducing the practical usability of protected images. We therefore introduce a PatchGAN-style discriminator $D_\phi$ to regularize local texture patterns and encourage the protected images to remain perceptually natural~\cite{isola2017image}.
			
%			Let
%			$D_\phi:[0,1]^{H\times W\times3}
%			\rightarrow\mathbb{R}$
%			denote the PatchGAN discriminator parameterized by $\phi$.
%			The output represents the realism score of the input image. 
			
			This adversarial regularization complements the norm constraint: the latter limits perturbation magnitude, while the discriminator constrains its spatial and perceptual distribution, thereby balancing attack effectiveness and visual fidelity.
			
			We adopt the Least Squares GAN objective for stable optimization. The discriminator loss is defined as:
			\begin{equation}
				\mathcal{L}_{D} =
				\frac{1}{2} \mathbb{E}_{x}
				\left[
				\left(D_{\phi}(x) - 1\right)^2
				\right]
				+
				\frac{1}{2} \mathbb{E}_{x_{adv}}
				\left[
				D_{\phi}(x_{adv})^2
				\right].
			\end{equation}
			The generator-side adversarial loss is:
			\begin{equation}
				\mathcal{L}_{adv} =
				\mathbb{E}_{x_{adv}}
				\left[
				\left(D_{\phi}(x_{adv}) - 1\right)^2
				\right].
			\end{equation}
			The final generator objective combines the MAD loss with the visual fidelity regularization:
			\begin{equation}
				\mathcal{L}_{G}
				=
				\mathcal{L}_{attn}
				+
				\lambda_{adv}\mathcal{L}_{adv},
			\end{equation}
			where $\lambda_{adv}$ controls the trade-off between adversarial disruption and visual fidelity.
			
			During training, $D_{\phi}$ and $G_{\theta}$ are optimized alternately. The discriminator learns to distinguish clean images from protected images, while the generator learns to both fool the discriminator and disrupt the multimodal attention features of the frozen I2V model. 
			After training converges, the discriminator and target I2V model are removed, while the trained $G_{\theta}$ and the frozen $E_{\mathrm{txt}}$ are retained for online image protection.
			
			\section{Experiments}
	
			\begin{table*}[t]
				\centering
%				\caption{Average comparative experimental results on CelebV-Text and UCF101.}
				\caption{Average comparative experimental results on CelebV-Text and UCF101. ``-'' denotes not applicable.}
				\label{tab:main_results}
				\resizebox{\textwidth}{!}{
					\begin{tabular}{l cc cccc c cccc}
						\toprule
						\multirow{2}{*}{Method}
						& \multirow{2}{*}{VRAM (GB)}
						& \multirow{2}{*}{TFLOPs}
						& \multicolumn{4}{c}{VBench}
						& \multicolumn{1}{c}{Q-Align}
						& \multicolumn{4}{c}{Gemini-3.1-flash-lite} \\
						\cmidrule(lr){4-7}
						\cmidrule(lr){8-8}
						\cmidrule(lr){9-12}
						& & &
						Sub. Cons. & Bg. Cons. & Mot. Smooth & Img. Quality &
						Vis. Score & Prompt Cons. & Temp. Cons. & Mot. Plaus. &
						Frame Qual. \\
						\midrule
						
						\multicolumn{12}{c}{\textbf{CogVideoX-5B}} \\
						\midrule
						Clean
						& - & -
						& 0.9105 & 0.9343 & 0.9786 & 0.4981
						& 0.5409 & 4.9400 & 3.9700 & 4.8000 & 3.7800 \\
						
						Random Noise
						& - & -
						& 0.9093 & 0.9329 & 0.9782 & 0.4885
						& 0.5299 & 4.9400 & 3.9300 & 4.7900 & 3.7000 \\
						
						PhotoGuard
						& 22.42 & 1438.02
						& 0.9016 & 0.9253 & 0.9778 & \textbf{0.4865}
						& \textbf{0.5171} & 4.8900 & \textbf{3.8400}
						& \textbf{4.6900} & \textbf{3.6400} \\
						
						\rowcolor[gray]{0.9}
						I2VShield (Ours)
						& \textbf{9.49} & \textbf{2.23}
						& \textbf{0.8962} & \textbf{0.9248} & \textbf{0.9756}
						& 0.4978 & 0.5216 & \textbf{4.8700}
						& 3.9200 & 4.7800 & 3.7450 \\
						
						\midrule
						\multicolumn{12}{c}{\textbf{Wan2.1-14B}} \\
						\midrule
						Clean
						& - & -
						& 0.8781 & 0.9195 & 0.9673 & 0.5224
						& 0.5196 & \textbf{4.8600} & 3.9300
						& 4.8000 & 3.6100 \\
						
						Random Noise
						& - & -
						& 0.8653 & 0.9170 & 0.9683 & 0.5281
						& 0.5186 & 4.9400 & 3.9600 & 4.8700 & 3.6500 \\
						
						PhotoGuard
						& 40.92 & 2133.51
						& 0.8379 & 0.9013 & 0.9601 & 0.4957
						& 0.4372 & 4.9200 & 3.8600 & 4.7500 & 3.5200 \\
						
						\rowcolor[gray]{0.9}
						I2VShield (Ours)
						& \textbf{12.27} & \textbf{4.95}
						& \textbf{0.8292} & \textbf{0.8968} & \textbf{0.9556}
						& \textbf{0.4895} & \textbf{0.4304} & 4.9300
						& \textbf{3.8300} & \textbf{4.6800} & \textbf{3.3600} \\
						
						\midrule
						\multicolumn{12}{c}{\textbf{OpenSora-V2-11B}} \\
						\midrule
						Clean
						& - & -
						& 0.9237 & 0.9495 & 0.9905 & 0.5428
						& 0.5242 & 4.8800 & 3.9800 & 4.8600 & 3.7400 \\
						
						Random Noise
						& - & -
						& 0.9241 & 0.9498 & 0.9896 & 0.5396
						& 0.5205 & 4.8800 & \textbf{3.9600}
						& 4.8800 & 3.7100 \\
						
						PhotoGuard
						& 34.73 & 377.50
						& 0.9267 & 0.9496 & 0.9888 & \textbf{0.5343}
						& \textbf{0.4901} & 4.9000 & 4.0300
						& 4.8500 & \textbf{3.6700} \\
						
						\rowcolor[gray]{0.9}
						I2VShield (Ours)
						& \textbf{9.79} & \textbf{14.59}
						& \textbf{0.9209} & \textbf{0.9474} & \textbf{0.9883}
						& 0.5434 & 0.5033 & \textbf{4.8500}
						& 4.0100 & \textbf{4.8000} & \textbf{3.6700} \\
						
						\bottomrule
					\end{tabular}
				}
			\end{table*}
			
			\subsection{Experimental Setup}
			
			\subsubsection{Datasets}
			We evaluate I2VShield on two datasets: UCF101~\cite{soomro2012ucf101} for complex human action generation, and CelebV-Text~\cite{yu2023celebv} for facial portrait animation. For both datasets, we randomly sample 900, 100, and 50 instances for training, validation, and testing, respectively.
			
			\subsubsection{Target Models and Baselines}
			We assess the defensive capability of I2VShield against three state-of-the-art DiT-based I2V models: CogVideoX-5B~\cite{hong2022cogvideo}, OpenSora-V2-11B~\cite{zheng2024open}, and Wan2.1-14B~\cite{wan2025wan}. Consistent with our threat model, we assume a white-box setting strictly during the offline training phase. We compare I2VShield against three baselines: (1) \textit{Clean} (original, unperturbed images); (2) \textit{Random Noise} (uniformly distributed noise under the same $\ell_\infty$ budget); and (3) \textit{PhotoGuard}, a representative gradient-based defense.
			
			\subsubsection{Evaluation Metrics}
%			As severe generation degradation characterizes effective attacks, lower scores across our metrics indicate a stronger defense. 
			Because more effective protection should cause greater degradation in unauthorized generations, lower scores generally indicate stronger protection.
			We employ VBench~\cite{huang2024vbench} to objectively assess spatial fidelity and temporal coherence (specifically subject/background consistency, motion smoothness, and image quality); Q-Align~\cite{wu2023q} for blind visual quality assessment; and a Vision-Language Model (VLM)\footnote{We use Gemini-3.1-flash-lite as an automated judge for scalable, zero-shot video assessment.} to evaluate semantic dimensions, including prompt consistency, temporal consistency, motion plausibility, and frame quality.

			\subsection{Main Results}
            \begin{figure*}[t] 
				\centering
				\includegraphics[width=\linewidth]{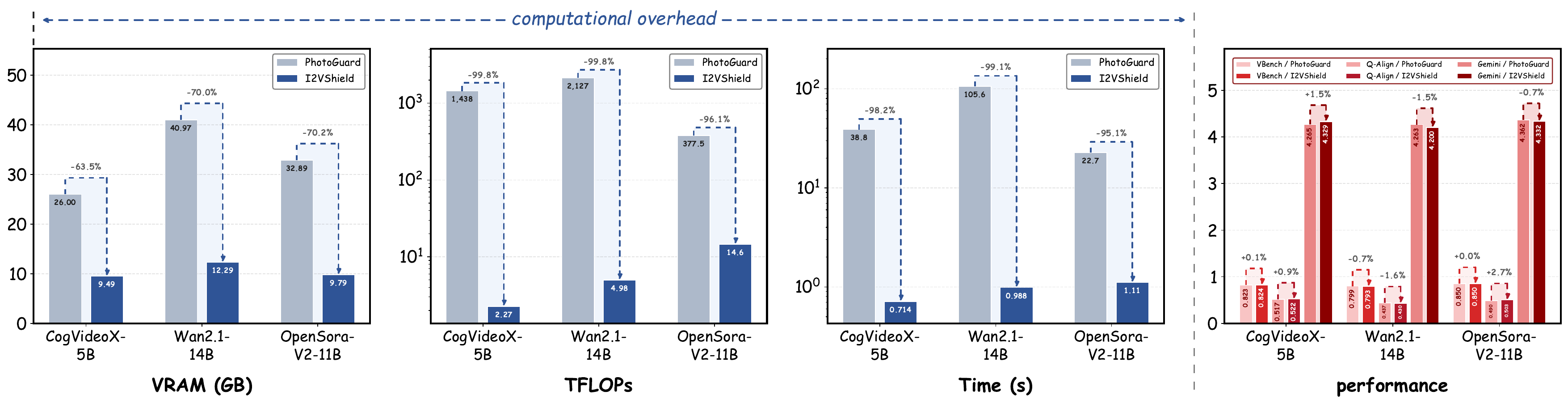}
%				\caption{Comparison of the computational overhead and performance of PhotoGuard and I2VShield on CelebV-Text.}
				\caption{Comparison of the computational overhead and performance of PhotoGuard and I2VShield. The proposed I2VShield significantly reduces computational overhead while achieving defense performance comparable to the baseline.}

				\label{computation}
			\end{figure*}
            
			\begin{figure*}[t] 
				\centering
				\includegraphics[width=\linewidth]{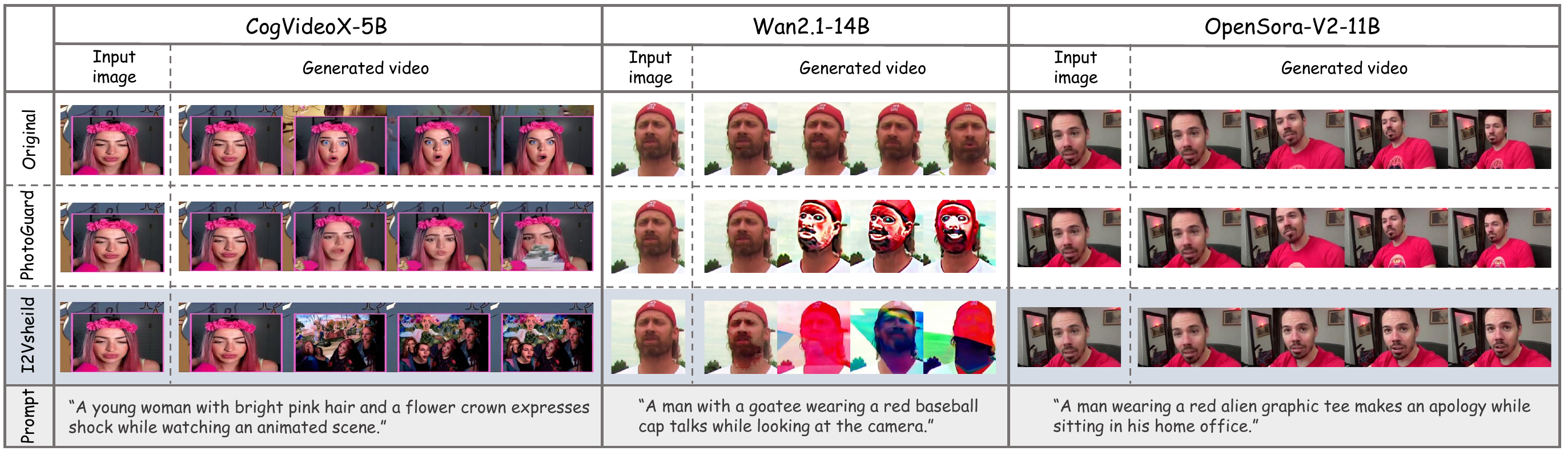}
				\caption{Qualitative comparison between the proposed method and the baseline. Our I2VShield produces more severe degradation in both visual appearance and temporal consistency.
				}
				\label{Qualitative}
			\end{figure*}
		
			\subsubsection{Quantitative Results}

			As shown in Table~\ref{tab:main_results}, when averaged across CelebV-Text and UCF101, I2VShield provides highly competitive protection across all evaluated DiT-based I2V models. Compared with PhotoGuard, I2VShield consistently achieves lower Subject Consistency, Background Consistency, and Motion Smoothness scores on CogVideoX-5B, Wan2.1-14B, and OpenSora-V2-11B, demonstrating stronger disruption of the generated videos' spatiotemporal coherence. The Gemini-based VLM evaluation reveals complementary, model-dependent effects: I2VShield obtains lower Prompt Consistency on CogVideoX-5B and OpenSora-V2-11B, while achieving lower Motion Plausibility on Wan2.1-14B and OpenSora-V2-11B. Moreover, I2VShield achieves these competitive protection effects with substantially lower VRAM consumption and computational cost than PhotoGuard across all three models, highlighting its efficiency and practical scalability.

			Figure~\ref{computation} further compares the per-image online protection costs and performance of I2VShield and PhotoGuard. The reported results exclude the offline training cost of I2VShield. We train a separate perturbation generator for each target model to accommodate its input resolution and text-conditioning interface, resulting in different online costs across models. PhotoGuard requires iterative forward and backward passes through the target I2V model for every input image, whereas I2VShield requires only text encoding and a single forward pass through the corresponding perturbation generator. Consequently, I2VShield substantially reduces peak GPU memory consumption, floating-point operations, and protection time across the evaluated settings. These results demonstrate the deployment-time efficiency of I2VShield after the perturbation generator has been trained.

			\subsubsection{Qualitative Results}

            Figure~\ref{Qualitative} presents a qualitative comparison of videos generated from clean reference images and images protected by PhotoGuard or I2VShield. When conditioned on clean images, all target I2V models consistently synthesize temporally coherent videos while preserving high subject fidelity.

            Multimodal Attention Disruption (MAD), employed by I2VShield, disrupts frame-level appearance and the temporal propagation of identity- and prompt-relevant cues. Consequently, all three models exhibit subject degradation and temporal inconsistencies, although manifestations vary across architectures. For CogVideoX-5B, the sequence progressively deviates from the reference subject and drifts toward unrelated content. For Wan2.1-14B, facial structures become distorted and color fidelity deteriorates across frames, accompanied by abrupt inter-frame variations. For OpenSora-V2-11B, facial motion is markedly suppressed, resulting in repetitive frames with limited semantic progression. In comparison, PhotoGuard only partially disrupts the generated content: although localized facial artifacts emerge, the identity, background, and motion trajectory remain recognizable in several cases. The consistent failure patterns induced by MAD across all three architectures demonstrate that it generalizes beyond model-specific artifact patterns and provides similarly effective protection against the evaluated I2V models. Additional qualitative examples are provided in the supplementary material.

			\subsection{Ablation Study}
			We evaluate three key designs of I2VShield on CelebV-Text:
			(1) text-adaptive perturbation generation, which incorporates the text embedding as an additional input to the perturbation generator,
			(2) the optimization target (our MAD attack vs. standard predicted noise attack), and
			(3) the PatchGAN discriminator ($\mathcal{L}_{adv}$) for fidelity regularization.
            
			\begin{table}[t]
				\centering
				\caption{Ablation results of text-conditioned perturbation generation.}
				\label{tab:ablation_text_embedding}
				\resizebox{\columnwidth}{!}{
					\begin{tabular}{cc ccc}
						\toprule
						\multicolumn{2}{c}{Perturbation Generator Input}
						& \multicolumn{3}{c}{Video Quality Degradation ($\downarrow$)} \\
						\cmidrule{1-2} \cmidrule{3-5}
						Visual Feature & Text Embedding
						& VBench-IQ & Q-Align & Gemini-MP \\
						\midrule
						\checkmark & 
						& {0.5755} & \textbf{0.5524} & {4.6600} \\
						\rowcolor[gray]{0.9}
						\checkmark & \checkmark
						& \textbf{0.5746} &  {0.5585} &  \textbf{4.5200} \\
						\bottomrule
					\end{tabular}
				}
			\end{table}
            
\subsubsection{Effectiveness of Text-adaptive Perturbation Generation}
As shown in Table~\ref{tab:ablation_text_embedding}, VBench-IQ and Q-Align measure visual quality using VBench and Q-Align, respectively, while Gemini-MP evaluates motion smoothness with Gemini; lower scores indicate stronger degradation. Adding text embeddings reduces VBench-IQ and Gemini-MP, demonstrating improved degradation of both visual quality and motion smoothness. Although Q-Align slightly increases slightly, the two variants remain broadly comparable on this metric. Overall, text conditioning provides complementary semantic guidance for generating more effective prompt-aware perturbations.

			\begin{table}[t]
				\centering
				\caption{Ablation results of attention disruption}
				\label{tab:ablation_main}
				\resizebox{\columnwidth}{!}{
					\begin{tabular}{c ccc}
						\toprule
						\multirow{2}{*}{Attack Target}
						& \multicolumn{3}{c}{Video Quality Degradation ($\downarrow$)} \\
						\cmidrule{2-4}
						& VBench-IQ & Q-Align & Gemini-MP \\
						\midrule
						Predicted Noise
						& 0.5749 & 0.5677 & 4.5600 \\
						\rowcolor[gray]{0.9}
						Attention Disruption
						& \textbf{0.5746} & \textbf{0.5585} & \textbf{4.5200} \\
						\bottomrule
					\end{tabular}
				}
			\end{table}
			
\begin{table}[t]
	\centering
	\caption{Fidelity comparison of protected reference images.}
	\label{tab:ablation_fidelity}
	\resizebox{\columnwidth}{!}{
		\begin{tabular}{c ccc}
			\toprule
			\multirow{2}{*}{Loss Function}
			& \multicolumn{3}{c}{Image Fidelity Metrics} \\
			\cmidrule{2-4}
			& LPIPS ($\downarrow$)
			& PSNR ($\uparrow$)
			& SSIM ($\uparrow$) \\
			\midrule
			w/o $\mathcal{L}_{adv}$
			& 0.3278
			& 30.5159
			& 0.6814 \\
			\rowcolor[gray]{0.9}
			w/ $\mathcal{L}_{adv}$
			& \textbf{0.3191}
			& \textbf{32.7229}
			& \textbf{0.7885} \\
			\bottomrule
		\end{tabular}
	}
\end{table}

            \subsubsection{Effectiveness of MAD Attack}
            As demonstrated in Table~\ref{tab:ablation_main}, targeting internal cross-attention features achieves more effective video degradation than perturbing the predicted noise space. Our full framework consistently outperforms the predicted noise attack across all evaluation metrics, with a particularly notable improvement on Q-Align, whose visual-quality score decreases from 0.5677 to 0.5585. These results suggest that perturbing cross-attention features more effectively disrupts multimodal feature alignment, leading to accumulated distortions in both visual appearance and temporal consistency throughout the video generation process.

			\subsubsection{Contribution of Discriminator Regularization}
\begin{figure}[!htbp]
				\centering
				\includegraphics[width=\linewidth]{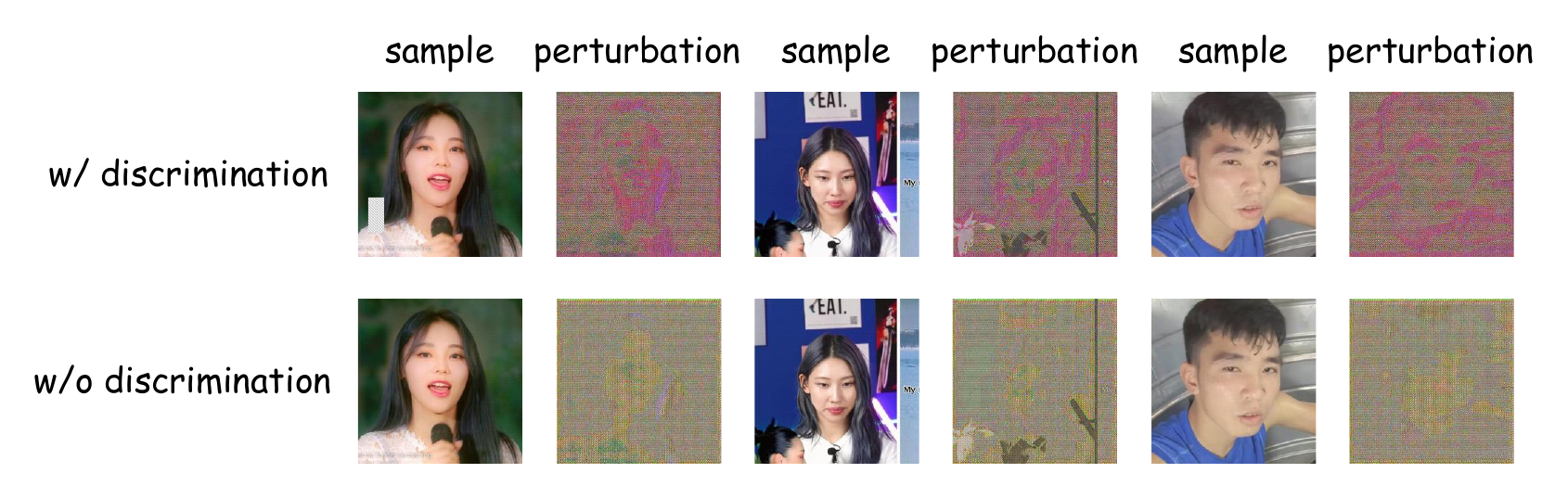}
				\caption{{Effect of discriminator regularization on protected image fidelity.} 
					%		The PatchGAN loss $\mathcal{L}_{adv}$ suppresses visible artifacts and encourages more natural, texture-adaptive perturbations.
				}
				\label{fig:ablation_fidelity}
			\end{figure}

			Table~\ref{tab:ablation_fidelity} and Figure~\ref{fig:ablation_fidelity} illustrate the impact of the PatchGAN discriminator. Without $\mathcal{L}_{adv}$, the MAD-only variant introduces noticeable structural and high-frequency artifacts. Incorporating $\mathcal{L}_{adv}$ improves PSNR by 7.2\%, reduces the LPIPS score by 2.7\%, and yields a more pronounced 15.7\% gain in SSIM, indicating that the discriminator is particularly effective in preserving structural information. These improvements suggest that adversarial supervision regularizes perturbation generation, encouraging texture-adaptive perturbations that better align with the underlying image distribution while maintaining the protection capability.
			
			\section{Conclusions}
			We propose I2VShield, an efficient proactive privacy defense against DiT-based I2V models. Unlike costly iterative gradient-based methods, I2VShield uses a lightweight generative framework to produce text-adaptive adversarial perturbations in a single forward pass. Its untargeted MAD attack maximizes feature deviation within DiT cross-attention layers, weakening image-text conditioning. Experiments across multiple I2V models demonstrate competitive protection performance with substantially lower inference time and memory consumption, supporting practical large-scale protection.

			\bibliographystyle{plainnat}
			\bibliography{ref}

@article{hong2022cogvideo,
  title={Cogvideo: Large-scale pretraining for text-to-video generation via transformers},
  author={Hong, Wenyi and Ding, Ming and Zheng, Wendi and Liu, Xinghan and Tang, Jie},
  journal={arXiv preprint arXiv:2205.15868},
  year={2022}
}

@article{soomro2012ucf101,
  title={Ucf101: A dataset of 101 human actions classes from videos in the wild},
  author={Soomro, Khurram and Zamir, Amir Roshan and Shah, Mubarak},
  journal={arXiv preprint arXiv:1212.0402},
  year={2012}
}

@inproceedings{yu2023celebv,
  title={Celebv-text: A large-scale facial text-video dataset},
  author={Yu, Jianhui and Zhu, Hao and Jiang, Liming and Loy, Chen Change and Cai, Weidong and Wu, Wayne},
  booktitle={Proceedings of the IEEE/CVF Conference on Computer Vision and Pattern Recognition},
  pages={14805--14814},
  year={2023}
}

@inproceedings{huang2024vbench,
  title={Vbench: Comprehensive benchmark suite for video generative models},
  author={Huang, Ziqi and He, Yinan and Yu, Jiashuo and Zhang, Fan and Si, Chenyang and Jiang, Yuming and Zhang, Yuanhan and Wu, Tianxing and Jin, Qingyang and Chanpaisit, Nattapol and others},
  booktitle={Proceedings of the IEEE/CVF Conference on Computer Vision and Pattern Recognition},
  pages={21807--21818},
  year={2024}
}

@article{wu2023q,
  title={Q-align: Teaching lmms for visual scoring via discrete text-defined levels},
  author={Wu, Haoning and Zhang, Zicheng and Zhang, Weixia and Chen, Chaofeng and Liao, Liang and Li, Chunyi and Gao, Yixuan and Wang, Annan and Zhang, Erli and Sun, Wenxiu and others},
  journal={arXiv preprint arXiv:2312.17090},
  year={2023}
}

@article{salman2023raising,
  title={Raising the cost of malicious ai-powered image editing},
  author={Salman, Hadi and Khaddaj, Alaa and Leclerc, Guillaume and Ilyas, Andrew and Madry, Aleksander},
  journal={arXiv preprint arXiv:2302.06588},
  year={2023}
}

@inproceedings{blattmann2023align,
  title={Align your latents: High-resolution video synthesis with latent diffusion models},
  author={Blattmann, Andreas and Rombach, Robin and Ling, Huan and Dockhorn, Tim and Kim, Seung Wook and Fidler, Sanja and Kreis, Karsten},
  booktitle={Proceedings of the IEEE/CVF conference on computer vision and pattern recognition},
  pages={22563--22575},
  year={2023}
}

@article{guo2023animatediff,
  title={Animatediff: Animate your personalized text-to-image diffusion models without specific tuning},
  author={Guo, Yuwei and Yang, Ceyuan and Rao, Anyi and Liang, Zhengyang and Wang, Yaohui and Qiao, Yu and Agrawala, Maneesh and Lin, Dahua and Dai, Bo},
  journal={arXiv preprint arXiv:2307.04725},
  year={2023}
}

@inproceedings{peebles2023scalable,
  title={Scalable diffusion models with transformers},
  author={Peebles, William and Xie, Saining},
  booktitle={Proceedings of the IEEE/CVF international conference on computer vision},
  pages={4195--4205},
  year={2023}
}

@article{brooks2024video,
  title={Video generation models as world simulators},
  author={Brooks, Tim and Peebles, Bill and Holmes, Connor and DePue, Will and Guo, Yufei and Jing, Leo and Schnurr, David and Taylor, Joe and Luhman, Troy and Luhman, Eric and others},
  journal={OpenAI Blog},
  volume={1},
  number={8},
  pages={1},
  year={2024}
}

@article{ma2024latte,
  title={Latte: Latent diffusion transformer for video generation},
  author={Ma, Xin and Wang, Yaohui and Chen, Xinyuan and Jia, Gengyun and Liu, Ziwei and Li, Yuan-Fang and Chen, Cunjian and Qiao, Yu},
  journal={arXiv preprint arXiv:2401.03048},
  year={2024}
}

@inproceedings{shan2023glaze,
  title={Glaze: Protecting artists from style mimicry by $\{$Text-to-Image$\}$ models},
  author={Shan, Shawn and Cryan, Jenna and Wenger, Emily and Zheng, Haitao and Hanocka, Rana and Zhao, Ben Y},
  booktitle={32nd USENIX Security Symposium (USENIX Security 23)},
  pages={2187--2204},
  year={2023}
}

@inproceedings{van2023anti,
  title={Anti-dreambooth: Protecting users from personalized text-to-image synthesis},
  author={Van Le, Thanh and Phung, Hao and Nguyen, Thuan Hoang and Dao, Quan and Tran, Ngoc N and Tran, Anh},
  booktitle={Proceedings of the IEEE/CVF International Conference on Computer Vision},
  pages={2116--2127},
  year={2023}
}

@article{kurakin2016adversarial,
  title={Adversarial machine learning at scale},
  author={Kurakin, Alexey and Goodfellow, Ian and Bengio, Samy},
  journal={arXiv preprint arXiv:1611.01236},
  year={2016}
}

@article{madry2017towards,
  title={Towards deep learning models resistant to adversarial attacks},
  author={Madry, Aleksander and Makelov, Aleksandar and Schmidt, Ludwig and Tsipras, Dimitris and Vladu, Adrian},
  journal={arXiv preprint arXiv:1706.06083},
  year={2017}
}

@inproceedings{isola2017image,
  title={Image-to-image translation with conditional adversarial networks},
  author={Isola, Phillip and Zhu, Jun-Yan and Zhou, Tinghui and Efros, Alexei A},
  booktitle={Proceedings of the IEEE conference on computer vision and pattern recognition},
  pages={1125--1134},
  year={2017}
}

@article{zheng2024open,
  title={Open-sora: Democratizing efficient video production for all},
  author={Zheng, Zangwei and Peng, Xiangyu and Yang, Tianji and Shen, Chenhui and Li, Shenggui and Liu, Hongxin and Zhou, Yukun and Li, Tianyi and You, Yang},
  journal={arXiv preprint arXiv:2412.20404},
  year={2024}
}

@article{wan2025wan,
  title={Wan: Open and advanced large-scale video generative models},
  author={Wan, Team and Wang, Ang and Ai, Baole and Wen, Bin and Mao, Chaojie and Xie, Chen-Wei and Chen, Di and Yu, Feiwu and Zhao, Haiming and Yang, Jianxiao and others},
  journal={arXiv preprint arXiv:2503.20314},
  year={2025}
}

@inproceedings{tulyakov2018mocogan,
  title={Mocogan: Decomposing motion and content for video generation},
  author={Tulyakov, Sergey and Liu, Ming-Yu and Yang, Xiaodong and Kautz, Jan},
  booktitle={Proceedings of the IEEE conference on computer vision and pattern recognition},
  pages={1526--1535},
  year={2018}
}

@article{yan2021videogpt,
  title={Videogpt: Video generation using vq-vae and transformers},
  author={Yan, Wilson and Zhang, Yunzhi and Abbeel, Pieter and Srinivas, Aravind},
  journal={arXiv preprint arXiv:2104.10157},
  year={2021}
}

@article{ho2020denoising,
  title={Denoising diffusion probabilistic models},
  author={Ho, Jonathan and Jain, Ajay and Abbeel, Pieter},
  journal={Advances in neural information processing systems},
  volume={33},
  pages={6840--6851},
  year={2020}
}

@inproceedings{rombach2022high,
  title={High-resolution image synthesis with latent diffusion models},
  author={Rombach, Robin and Blattmann, Andreas and Lorenz, Dominik and Esser, Patrick and Ommer, Bj{\"o}rn},
  booktitle={Proceedings of the IEEE/CVF conference on computer vision and pattern recognition},
  pages={10684--10695},
  year={2022}
}

@inproceedings{rossler2019faceforensics++,
  title={Faceforensics++: Learning to detect manipulated facial images},
  author={Rossler, Andreas and Cozzolino, Davide and Verdoliva, Luisa and Riess, Christian and Thies, Justus and Nie{\ss}ner, Matthias},
  booktitle={Proceedings of the IEEE/CVF international conference on computer vision},
  pages={1--11},
  year={2019}
}

@inproceedings{wang2020cnn,
  title={CNN-generated images are surprisingly easy to spot... for now},
  author={Wang, Sheng-Yu and Wang, Oliver and Zhang, Richard and Owens, Andrew and Efros, Alexei A},
  booktitle={Proceedings of the IEEE/CVF conference on computer vision and pattern recognition},
  pages={8695--8704},
  year={2020}
}

@article{dosovitskiy2020image,
  title={An image is worth 16x16 words: Transformers for image recognition at scale},
  author={Dosovitskiy, Alexey and Beyer, Lucas and Kolesnikov, Alexander and Weissenborn, Dirk and Zhai, Xiaohua and Unterthiner, Thomas and Dehghani, Mostafa and Minderer, Matthias and Heigold, Georg and Gelly, Sylvain and others},
  journal={arXiv preprint arXiv:2010.11929},
  year={2020}
}

@inproceedings{xu2025multimodal,
  title={A multimodal deviation perceiving framework for weakly-supervised temporal forgery localization},
  author={Xu, Wenbo and Wu, Junyan and Lu, Wei and Luo, Xiangyang and Wang, Qian},
  booktitle={Proceedings of the 33rd ACM International Conference on Multimedia},
  pages={11581--11589},
  year={2025}
}

@article{qu2025idguard,
  author={Qu, Zuomin and Lu, Wei and Luo, Xiangyang and Wang, Qian and Cao, Xiaochun},
  journal={IEEE Transactions on Pattern Analysis and Machine Intelligence}, 
  title={ID-Guard: A Universal Framework for Combating Facial Manipulation via Breaking Identification}, 
  year={2026},
  volume={48},
  number={2},
  pages={1720-1735}
}

@inproceedings{gui2025i2vguard,
  title={I2vguard: Safeguarding images against misuse in diffusion-based image-to-video models},
  author={Gui, Dongnan and Guo, Xun and Zhou, Wengang and Lu, Yan},
  booktitle={Proceedings of the Computer Vision and Pattern Recognition Conference},
  pages={12595--12604},
  year={2025}
}

			\clearpage
			\onecolumn
			\appendix
			\setcounter{secnumdepth}{1}

			\begin{center}
				{\Large\bfseries Supplementary Material}
			\end{center}

			\section{Quantitative Results Details}

			The main paper reports the average quantitative results of the proposed I2VShield and the baseline method across the CelebV-Text and UCF101 datasets. Here, we further provide the quantitative results on each dataset separately for a more detailed comparison.

			\begin{table*}[!htbp]
				\centering
				\caption{Comparative Experimental Results on CelebV-Text}
				\resizebox{\textwidth}{!}{
					\begin{tabular}{l cc cccc c cccc}
						\toprule
						\multirow{2}{*}{Method} & \multirow{2}{*}{VRAM(GB)} & \multirow{2}{*}{TFLOPs} & \multicolumn{4}{c}{VBench} & \multicolumn{1}{c}{Q-align} & \multicolumn{4}{c}{Gemini-3.1-flash-lite} \\
						\cmidrule(lr){4-7} \cmidrule(lr){8-8} \cmidrule(lr){9-12}
						& & & Sub. Cons. & Bg. Cons. & Mot. Smooth & Img. Quality & Vis. Score & Prompt Cons. & Temp. Cons. & Mot. Plaus. & Frame Qual. \\
						\midrule
						\multicolumn{12}{c}{\textbf{CogVideoX-5B}} \\
						\midrule
						Clean & - & - & 0.9195 & 0.9383 & 0.9767 & 0.6092 & 0.7118 & 4.8800 & 3.8800 & 4.6600 & 3.8200 \\
						Random Noise & - & - & 0.9192 & 0.9357 & 0.9751 & 0.6042 & 0.7045 & 4.8800 & 3.8600 & 4.6600 & 3.8000 \\
						PhotoGuard & 26.00  & 1438.02  & 0.9060 & 0.9236 & 0.9750 & \textbf{0.5840} & {0.6590} & 4.7800 & \textbf{3.7200} & \textbf{4.4400} & \textbf{3.6600} \\
						\rowcolor[gray]{0.9} I2VShield (Ours) & \textbf{9.49}  &  \textbf{2.27}  & \textbf{0.9004} & \textbf{0.9182} & \textbf{0.9721} & 0.6092 & \textbf{0.6585} & \textbf{4.7400} & 3.8200 & 4.6800 & 3.7200 \\
						\midrule
						\multicolumn{12}{c}{\textbf{Wan2.1-14B}} \\
						\midrule
						Clean & - & - & 0.8877 & 0.9202 & 0.9712 & 0.6163 & 0.6998 & \textbf{4.7800} & 3.8800 & 4.6800 & 3.8200 \\
						Random Noise & - & - & 0.8861 & 0.9218 & 0.9695 & 0.6247 & 0.7089 & 4.8800 & 3.9200 & 4.7600 & 3.9400 \\
						PhotoGuard & 40.97  & 2127.09  & 0.8467 & 0.8962 & 0.9627 & 0.5834 & 0.5723 & 4.8800 & \textbf{3.7200} & 4.5400 & 3.6600 \\
						\rowcolor[gray]{0.9} I2VShield (Ours) & \textbf{12.29} &\textbf{4.98} & \textbf{0.8303} & \textbf{0.8864} & \textbf{0.9594} & \textbf{0.5746} & \textbf{0.5585} & 4.8600 & 3.7400 & \textbf{4.5200} & \textbf{3.6000} \\
						\midrule
						\multicolumn{12}{c}{\textbf{OpenSora-V2-11B}} \\
						\midrule
						Clean & - & - & 0.9266 & 0.9504 & 0.9913 & 0.5981 & 0.6412 & 4.8200 & 3.9800 & 4.7800 & 3.8800 \\
						Random Noise & - & - & 0.9296 & 0.9493 & 0.9907 & 0.6002 & 0.6311 & 4.7600 & \textbf{3.9000} & 4.7600 & \textbf{3.8000} \\
						PhotoGuard & 32.89 & 377.50  & {0.9297} & {0.9491} & 0.9895 & \textbf{0.5969} & \textbf{0.5946} & 4.8000 & 4.0200 & 4.7200 & 3.8400 \\
						\rowcolor[gray]{0.9} I2VShield (Ours) & \textbf{9.78} & \textbf{14.59} & \textbf{0.9223} & \textbf{0.9476} & \textbf{0.9894} & 0.6098 & 0.6287 & \textbf{4.7000} & 3.9600 & \textbf{4.7000} & 3.8800 \\
						\bottomrule
					\end{tabular}
				}
			\end{table*}

			\begin{table*}[!htbp]
				\centering
				\caption{Comparative Experimental Results on UCF101}
				\resizebox{\textwidth}{!}{
					\begin{tabular}{l cc cccc c cccc}
						\toprule
						\multirow{2}{*}{Method} & \multirow{2}{*}{VRAM(GB)} & \multirow{2}{*}{TFLOPs} & \multicolumn{4}{c}{VBench} & \multicolumn{1}{c}{Q-align} & \multicolumn{4}{c}{Gemini-3.1-flash-lite} \\
						\cmidrule(lr){4-7} \cmidrule(lr){8-8} \cmidrule(lr){9-12}
						& & & Sub. Cons. & Bg. Cons. & Mot. Smooth & Img. Quality & Vis. Score & Prompt Cons. & Temp. Cons. & Mot. Plaus. & Frame Qual. \\
						\midrule
						\multicolumn{12}{c}{\textbf{CogVideoX-5B}} \\
						\midrule
						Clean & - & - & 0.9014 & 0.9302 & 0.9805 & 0.3869 & 0.3700 & \textbf{5.0000} & 4.0600 & 4.9400 & 3.7400 \\
						Random Noise & - & - & 0.8994 & 0.9300 & 0.9813 & \textbf{0.3728} & \textbf{0.3552} & \textbf{5.0000} & 4.0000 & {4.9200} & \textbf{3.6000} \\
						PhotoGuard & 18.85 & 1438.02 & 0.8972 & \textbf{0.9270} & 0.9805 & 0.3890 & 0.3752 & \textbf{5.0000} & \textbf{3.9600} & 4.9400 & 3.6200 \\
						\rowcolor[gray]{0.9} I2VShield (Ours) & \textbf{9.49} & \textbf{2.19} & \textbf{0.8920} & 0.9314 & \textbf{0.9791} & 0.3864 & 0.3847 & \textbf{5.0000} & 4.0200 & \textbf{4.8800} & 3.7700 \\
						\midrule
						\multicolumn{12}{c}{\textbf{Wan2.1-14B}} \\
						\midrule
						Clean & - & - & 0.8685 & 0.9188 & 0.9633 & 0.4284 & 0.3393 & \textbf{4.9400} & 3.9800 & 4.9200 & 3.4000 \\
						Random Noise & - & - & 0.8444 & 0.9121 & 0.9670 & 0.4314 & 0.3283 & 5.0000 & 4.0000 & 4.9800 & 3.3600 \\
						PhotoGuard & 40.86 & 2139.92 & 0.8291 & \textbf{0.9064} & 0.9574 & 0.4079 & \textbf{0.3021} & 4.9600 & 4.0000 & 4.9600 & 3.3800 \\
						\rowcolor[gray]{0.9} I2VShield (Ours)
						& \textbf{12.25} & \textbf{4.91}
						& \textbf{0.8281} & 0.9071 & \textbf{0.9518} & \textbf{0.4043} & 0.3022 & 5.0000 & \textbf{3.9200} & \textbf{4.8400} & \textbf{3.1200} \\
						\midrule
						\multicolumn{12}{c}{\textbf{OpenSora-V2-11B}} \\
						\midrule
						Clean & - & - & 0.9208 & 0.9486 & 0.9896 & 0.4875 & 0.4071 & \textbf{4.9400} & \textbf{3.9800} & {4.9400} & 3.6000 \\
						Random Noise & - & - & \textbf{0.9186} & 0.9502 & 0.9885 & 0.4789 & 0.4099 & 5.0000 & 4.0200 & 5.0000 & 3.6200 \\
						PhotoGuard & 36.56 & 377.50 & 0.9236 & 0.9501 & 0.9880 & \textbf{0.4717} & 0.3855 & 5.0000 & 4.0400 & 4.9800 & 3.5000 \\
						\rowcolor[gray]{0.9} I2VShield (Ours)
						& \textbf{9.79} & \textbf{14.59}
						& 0.9195 & \textbf{0.9472} & \textbf{0.9871} & 0.4769 & \textbf{0.3778} & 5.0000 & 4.0600 & \textbf{4.9000} & \textbf{3.4600} \\
						\bottomrule
					\end{tabular}
				}
			\end{table*}

			\section{More Qualitative Results}

\begin{figure}[t]
				\centering
				\includegraphics[width=\textwidth]{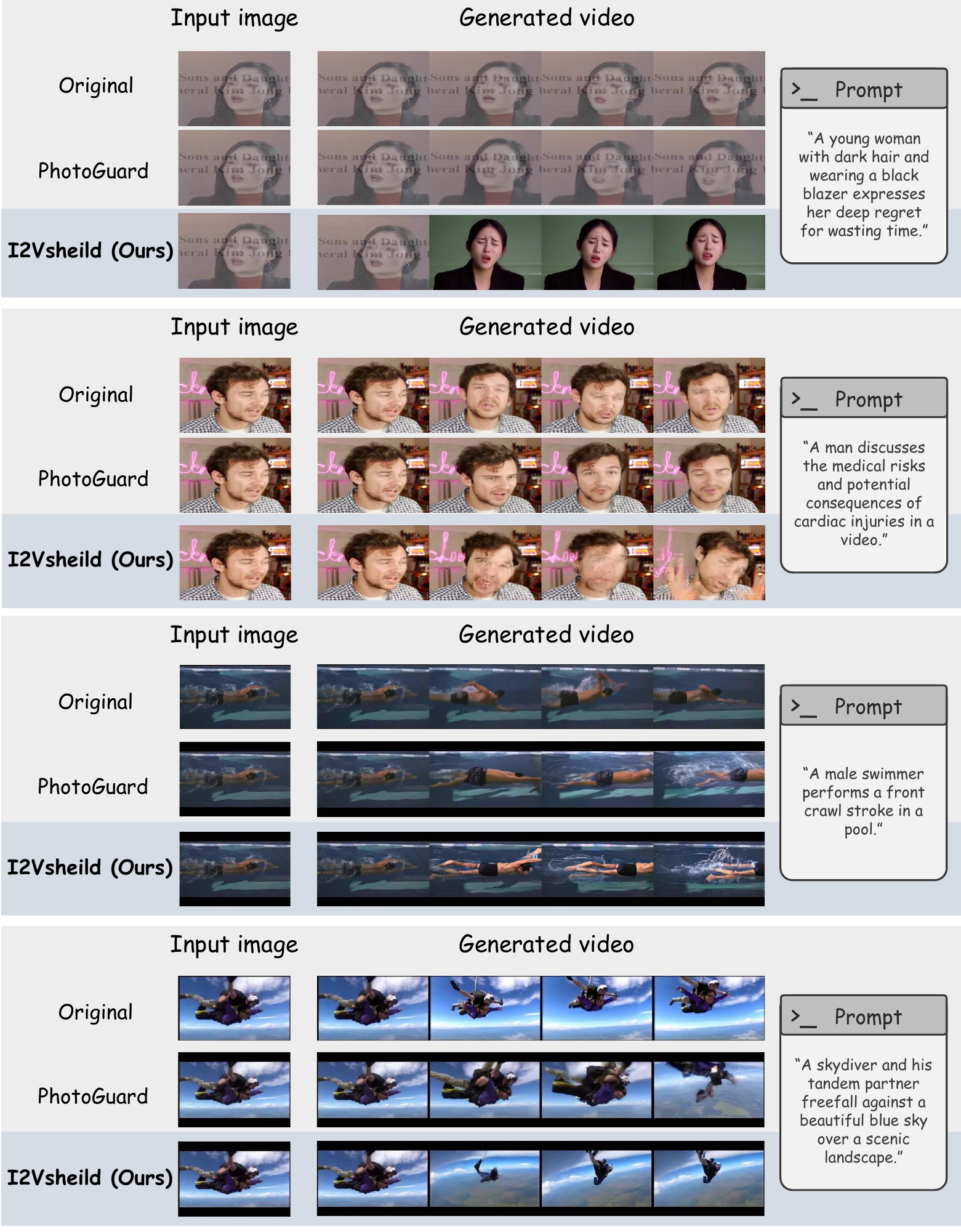}
				\caption{Qualitative comparison of I2VShield and baseline methods for protecting reference images against CogVideoX-5B.}
				\label{fig:CogVideoX-5B}
			\end{figure}

\begin{figure}[t]
				\centering
				\includegraphics[width=\textwidth]{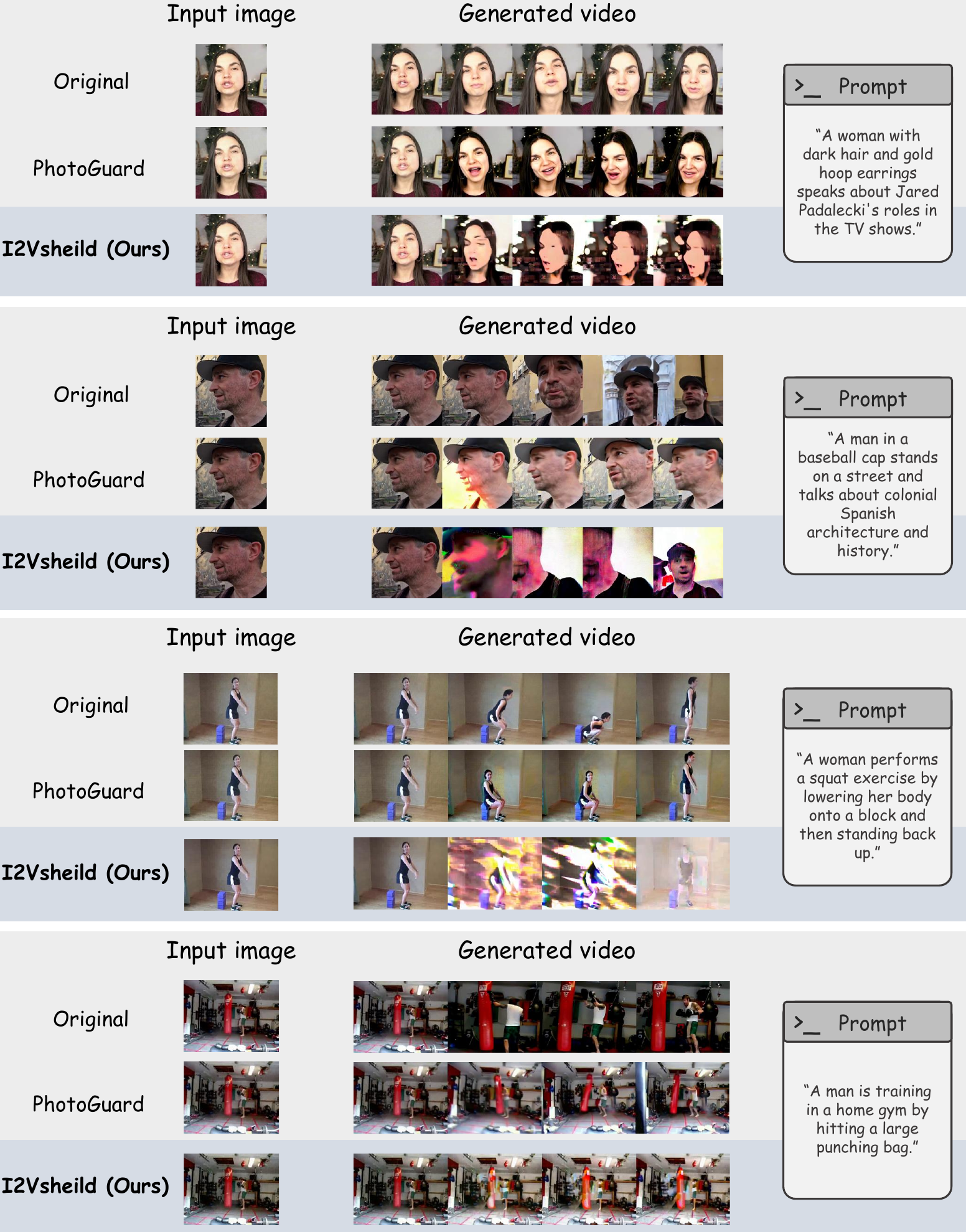}
				\caption{Qualitative comparison of I2VShield and baseline methods for protecting reference images against Wan2.1-14B.}
				\label{fig:wan}
			\end{figure}

\begin{figure}[t]
				\centering
				\includegraphics[width=\textwidth]{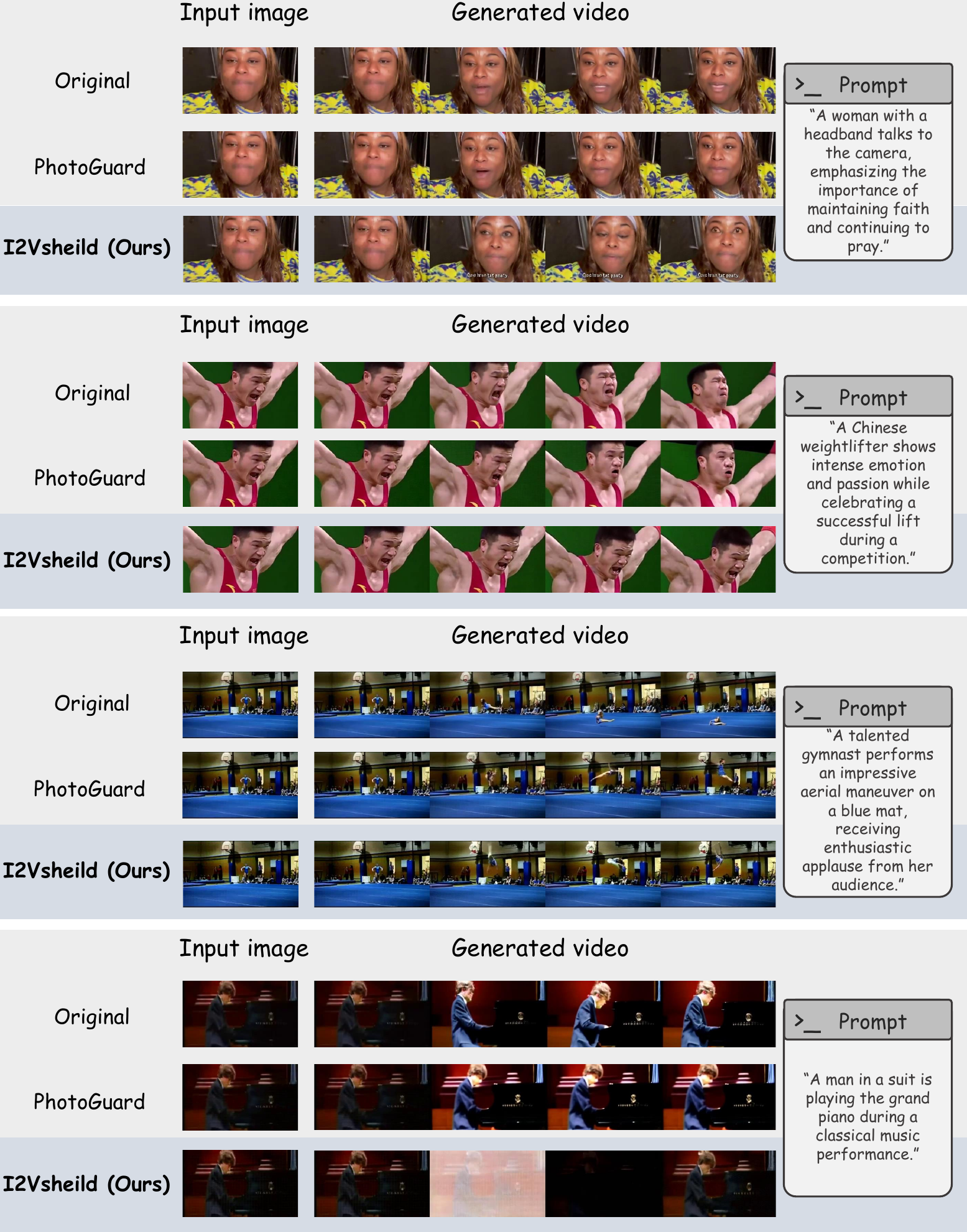}
				\caption{Qualitative comparison of I2VShield and baseline methods for protecting reference images against OpenSora-V2-11B.}
				\label{fig:opensora}
			\end{figure}

			We provide additional qualitative comparisons between the proposed I2VShield and the baseline method for defending against CogVideoX-5B, Wan2.1-14B, and OpenSora-V2-11B, respectively. Please refer to Figure \ref{fig:CogVideoX-5B}, \ref{fig:wan}, and \ref{fig:opensora}.

		\end{document}